\documentclass[letterpaper, 10 pt, conference]{ieeeconf}  

\IEEEoverridecommandlockouts                              

\usepackage{graphics} 
\usepackage{epsfig} 
\usepackage{mathptmx} 
\usepackage{times} 
\usepackage{amsmath} 
\usepackage{amssymb}  
\usepackage{cite}
\usepackage{algorithmic}
\usepackage{graphicx}
\usepackage{textcomp}
\usepackage{xcolor}
\usepackage{multirow}
\usepackage[flushleft]{threeparttable}

\title{Bayesian Approximation-Based Trajectory \\Prediction and Tracking with 4D Radar}

\author{Dong-In Kim$^{1}$, Dong-Hee Paek$^{2}$, Seung-Hyun Song$^{3}$, and Seung-Hyun~Kong$^{2*}$
\thanks{$^{*}$Corresponding author (e-mail: skong@kaist.ac.kr).}%
\thanks{$^{1}$D.-I. Kim was with the CCS Graduate School of Mobility, Korea Advanced Institute of Science and Technology, Daejeon, Korea, 34051, during this work. He is currently with Hyundai Motor Company, Republic of Korea (e-mail: dongin.kim@hyundai.com).}%
\thanks{$^{2}$D.-H. Paek and S.-H. Kong are with the CCS Graduate School of Mobility, Korea Advanced Institute of Science and Technology, Daejeon, Korea, 34051 (e-mail: \{donghee.paek, skong\}@kaist.ac.kr).}%
\thanks{$^{3}$S.-H. Song is with the Graduate School of Advanced Security and Technology, Korea Advanced Institute of Science and Technology, Daejeon, Korea, 34051 (e-mail: shyun@kaist.ac.kr).}%
}

\begin{document}

\maketitle
\thispagestyle{empty}
\pagestyle{empty}

\begin{abstract}
Accurate 3D multi-object tracking (MOT) is vital for autonomous vehicles, yet LiDAR and camera-based methods degrade in adverse weather. Meanwhile, Radar-based solutions remain robust but often suffer from limited vertical resolution and simplistic motion models. Existing Kalman filter–based approaches also rely on fixed noise covariance, hampering adaptability when objects make sudden maneuvers.
We propose \emph{Bayes-4DRTrack}, a 4D Radar-based MOT framework that adopts a transformer-based motion prediction network to capture nonlinear motion dynamics and employs Bayesian approximation in both detection and prediction steps. Moreover, our two-stage data association leverages Doppler measurements to better distinguish closely spaced targets. 
Evaluated on the K-Radar dataset (including adverse weather scenarios), Bayes-4DRTrack demonstrates a 5.7\% gain in Average Multi-Object Tracking Accuracy (AMOTA) over methods with traditional motion models and fixed noise covariance. These results showcase enhanced robustness and accuracy in demanding, real-world conditions.
\end{abstract}


\section{Introduction}
Accurate 3D multi-object tracking (MOT) systems are essential for autonomous vehicles, as they furnish the precise location, size, and velocity of surrounding objects, enabling efficient path planning and collision avoidance. Although LiDAR and camera-based systems have excelled in normal weather conditions, their performance deteriorates under adverse weather (e.g., fog, snow, or heavy rain) due to visibility issues and laser scattering \cite{adversetrack}. Conversely, Radar-based sensors maintain robust performance in such conditions owing to longer wavelengths and ability to penetrate rain and snow \cite{kradar}.

Conventional Radar-based systems, however, have been limited by lower vertical resolution, typically confining object detection to 2D bird’s-eye-view (BEV) or range–Doppler representations \cite{dopplertrack,ramaptrack}. Recent advances in multi-input multi-output (MIMO) antenna technology have led to the emergence of four-dimensional (4D) Radar, which provides measurements in range, azimuth, Doppler, and elevation \cite{kradar,tj4d,vod,enhanced}. This added dimension of height allows 4D Radar systems to more accurately capture 3D object positions.

3D MOT systems commonly follow a tracking-by-detection (TBD) pipeline: objects are independently detected in each frame and then matched over time. Although basic Kalman filter implementations with constant-velocity (CV) motion models have been widely used \cite{kalman}, they struggle in real-world scenarios where objects often undergo complex, nonlinear accelerations. Moreover, these systems typically employ fixed noise covariance in the state-correction step \cite{kalman}, which is insufficient for dealing with dynamic uncertainties such as abrupt changes in object motion. This shortfall can lead to a pronounced drop in overall tracking performance.

To address these limitations, we propose \emph{Bayes-4DRTrack}, a 4D Radar-based MOT system distinguished by two core innovations. First, it leverages a transformer-based motion prediction network \cite{transformer} to capture intricate temporal patterns and achieve more accurate trajectory predictions, especially in the presence of sudden velocity changes \cite{ab3dmot,probabilistic,simpletrack,score}. Second, the system applies Bayesian approximation techniques \cite{mcdropout,lossattenuation} to both detection and prediction steps, offering a dynamic approach to noise modeling. This framework replaces the conventional fixed noise covariance with an adaptive approach, better reflecting real-time uncertainties and enhancing system reliability.

\begin{figure*}[t]
  \centering
  \includegraphics[width=0.85\linewidth]{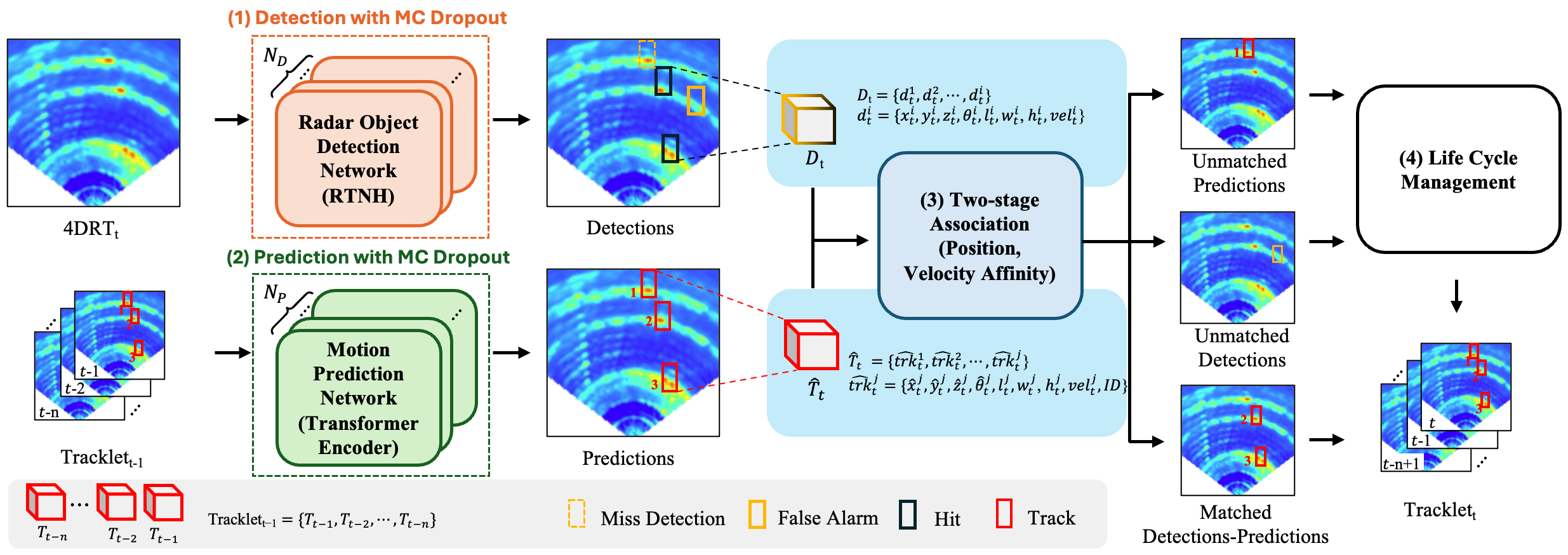}
  \caption{Structure of Bayes-4DRTrack. The proposed 3D MOT system consists of:
  (1) a Bayesian approximation–based 4D Radar detector, which produces object positions $\{d_t^i\}$;
  (2) a Bayesian approximation–based transformer motion prediction network, which outputs predicted tracks $\{\hat{trk}_t^j\}$;
  (3) a two-stage data association leveraging both Mahalanobis distance and Doppler velocity;
  and (4) a life-cycle management module for initializing/removing tracks.}
  \label{fig:architecture}
\end{figure*}

Our design further incorporates a two-stage data association approach that uses Doppler measurements, a distinguishing feature of Radar sensors. This velocity-based information effectively disambiguates objects with similar spatial characteristics but different relative velocities, mitigating errors in identity management and reducing ID switches. By integrating direct radial velocity measurements, we seek to improve the overall consistency and reliability of the tracking process in dense, fast-changing traffic scenarios.

In summary, our main contributions are as follows:
\begin{itemize}
    \item We propose \emph{Bayes-4DRTrack}, which, to our knowledge, is the first 4D Radar-based 3D MOT system to apply Bayesian approximation (via Monte Carlo (MC) Dropout and Loss Attenuation) to both detection and prediction stages. 
    \item We develop a two-stage data association method that exploits the unique Doppler information provided by 4D Radar, enhancing the system’s ability to distinguish true positives from false positives with similar spatial patterns.
    \item We demonstrate that Bayes-4DRTrack outperforms state-of-the-art 3D MOT systems on the K-Radar dataset, achieving a 5.7\% improvement in AMOTA. These gains underscore the advantages of incorporating dynamic uncertainty modeling and Doppler-based association in real-world autonomous driving applications.
\end{itemize}

The remainder of this paper is organized as follows: Section \ref{sec:sec2} reviews recent developments in 3D MOT and approaches to uncertainty estimation in detection and tracking. Section \ref{sec:sec3} outlines the technical details of Bayes-4DRTrack, including Bayesian approximation for both detection and prediction. Section \ref{sec:sec4} discusses our experimental results on the K-Radar dataset and compares our system with existing methods. Finally, Section \ref{sec:sec5} concludes this work and discusses future directions.

\section{Related Work}
\label{sec:sec2}
This section surveys current research trends in 3D MOT systems and uncertainty estimation strategies for object detection and tracking.

\subsection{3D Multi-Object Tracking Systems}
\label{sec:sec2a}

\subsubsection{LiDAR-based 3D MOT}
LiDAR-based 3D MOT systems typically follow a tracking-by-detection paradigm and utilize CV motion models with Kalman filtering \cite{ab3dmot,probabilistic,score,simpletrack}. Recent trends focus on seamlessly integrating velocity or confidence cues from the detector \cite{centerpoint}, as well as refining the association stage through spatiotemporal attention \cite{3dmot}. These enhancements improve overall accuracy but can suffer under adverse weather or sensor occlusion.

\subsubsection{Camera-based 3D MOT}
Camera-only 3D MOT systems rely heavily on dense RGB inputs and leverage various network architectures (e.g., LSTM-based motion models \cite{mono,cc3dt} and attention mechanisms \cite{mutr3d,spatio}). While such systems benefit from rich semantic details, their performance is limited in low-visibility or adverse weather conditions.

\subsubsection{LiDAR–Camera Fusion 3D MOT}
Fusion-based approaches combine LiDAR’s precise depth measurements with camera’s dense appearance cues, either in a tracking-by-detection pipeline \cite{gnn3dmot} or a joint detection and tracking framework \cite{jmodt,alphatrack}. While these systems often achieve superior performance under good conditions, they degrade significantly when either sensor becomes unreliable.

\subsubsection{4D Radar-based 3D MOT}
4D Radar technology offers a robust solution for tracking under challenging conditions due to its ability to provide 3D position and Doppler velocity, remaining relatively unaffected by fog, rain, or snow. While noise and resolution are still limiting factors compared to LiDAR, 4D Radar is quickly advancing toward more accurate 3D object detection and tracking \cite{kradar,tj4d}. This paper focuses on 4D Radar as the primary sensor, aiming to leverage its resilience under adverse conditions and direct velocity measurements.

\subsection{Uncertainty Estimation for Object Detection and Tracking}
\label{sec:sec2b}

\subsubsection{Types of Uncertainty}
Uncertainty in deep learning broadly falls into two categories: \emph{aleatoric uncertainty}, which originates from inherent sensor noise or measurement inconsistencies, and \emph{epistemic uncertainty}, stemming from model uncertainty or limited training data \cite{lossattenuation}.

\subsubsection{Uncertainty Estimation Methods}
Common strategies for estimating uncertainty include MC Dropout \cite{mcdropout}, Deep Ensembles \cite{deepensemble}, and Loss Attenuation \cite{lossattenuation}. 
MC Dropout repeatedly samples the network at inference by randomly dropping neurons, approximating the posterior distribution over network parameters. Deep Ensembles trains multiple independent models and aggregate their predictions. Loss Attenuation uses Gaussian-based modeling to dynamically weight the training loss according to estimated input or output uncertainty. These methods have been successfully employed in depth estimation \cite{depth}, semantic segmentation \cite{stochasticseg}, object detection \cite{lidaruncertainty,griduncertainty}, and visual tracking \cite{trackuncertainty}.

\subsubsection{Uncertainty in Detection and Tracking}
Although uncertainty estimation has proven beneficial in a variety of perception tasks \cite{reviewuncertainty}, relatively few 3D MOT systems integrate uncertainty estimates in both detection and prediction modules. For instance, \cite{lidaruncertainty} and \cite{griduncertainty} compute epistemic and aleatoric uncertainties in 3D detection tasks. Meanwhile, \cite{trackuncertainty} uses uncertainty to filter out low-confidence training samples in visual tracking. This paper tries to address the gap by applying Bayesian approximation in both detection and prediction for a 4D Radar-based 3D MOT system, aiming to improve robustness in uncertain environments.

\section{Proposed 3D Multi-Object Tracking System}
\label{sec:sec3}

We propose \emph{Bayes-4DRTrack}, a 4D Radar-based 3D MOT framework that integrates Bayesian approximation in both detection and motion prediction. Our system also incorporates a two-stage data association technique that leverages Doppler measurements to refine matching.

\subsection{Overview of Bayesian Approximation in Tracking}
\label{sec:sec3a}
Bayesian approximation improves system reliability by modeling uncertainties arising from noise, sensor degradation, or abrupt object maneuvers. Specifically, we apply MC Dropout \cite{mcdropout} and Loss Attenuation \cite{lossattenuation} in both detection and prediction. 

In detection, MC Dropout helps mitigate overfitting in a noisier environment (4D Radar has lower resolution than LiDAR). It produces an ensemble of object-detection outputs per frame, from which we take an average bounding-box estimation. Meanwhile, Loss Attenuation dynamically scales the training loss based on observed input noise, further improving the detector’s robustness. 

In prediction, MC Dropout again generates multiple trajectory estimates, and the mean and variance of these estimates serve as the predicted state and associated uncertainty. This enables more adaptive object tracking, especially during sudden velocity changes.

\subsection{Bayesian Approximation–Based Object Detection}
\label{sec:sec3b}
We employ the Radar Tensor Network with Height (RTNH) \cite{kradar} as our base 4D Radar detector. RTNH encodes 3D spatial information (X, Y, Z) via multiple layers of 3D sparse convolution, operating on a 4D Radar tensor (range, azimuth, Doppler, elevation).

\textbf{Loss Attenuation for Detection:}
Let $x_i$ denote an input sample, $f(x_i)$ the model’s prediction, and $y_i$ the ground-truth label. We incorporate an observation noise parameter $\sigma(x_i)$ that adjusts loss weights according to input uncertainty:
\begin{equation}
L_d = \frac{1}{N} \sum_{i=1}^{N} \left( \frac{1}{2\sigma(x_i)^2} \|y_i - f(x_i)\|^2 + \frac{1}{2} \log \sigma(x_i)^2 \right).
\end{equation}
Here, higher uncertainty $\sigma(x_i)$ reduces the loss weight, making the detector more resilient to noisy or ambiguous samples.

\begin{figure}[t]
  \centering
  \includegraphics[width=0.8\linewidth]{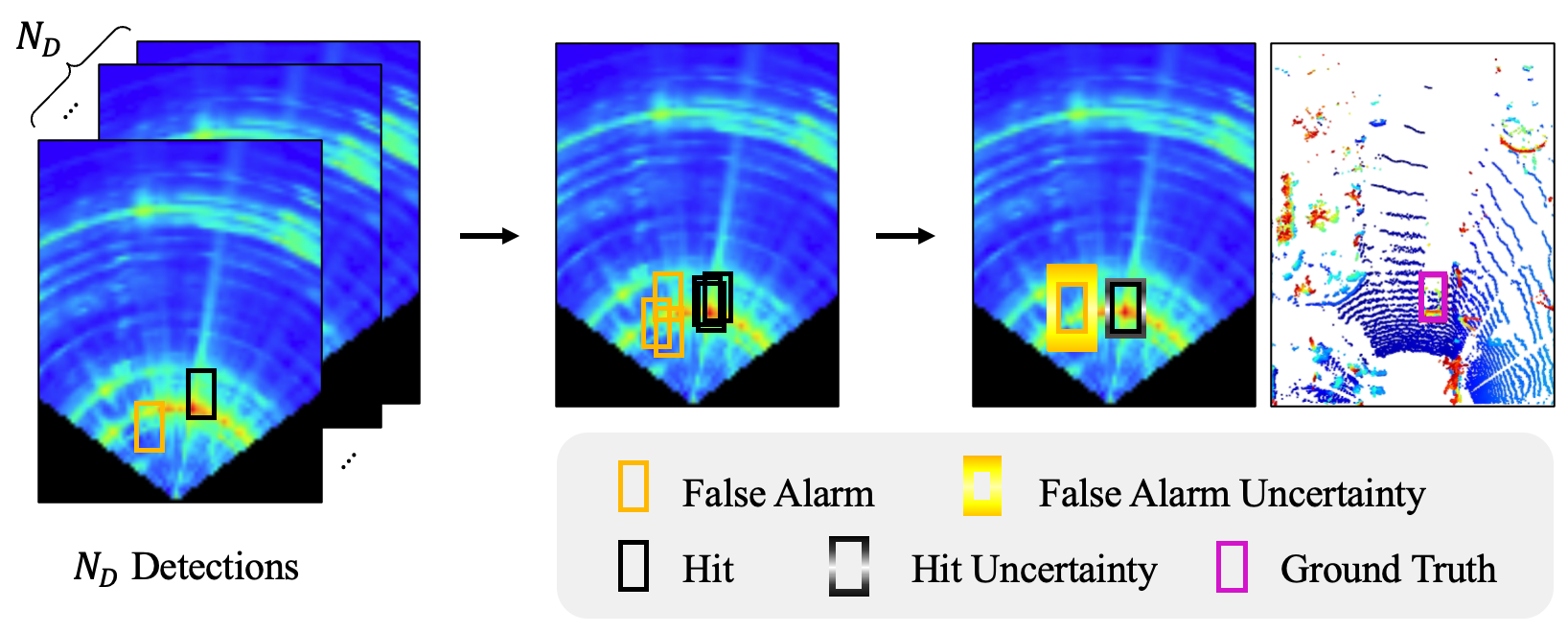}
  \caption{Bayesian approximation–based object detection results. 
  Using MC Dropout, $N_D$ detection outputs are generated per frame. 
  Bounding boxes are clustered via IoU; 
  within each cluster, box parameters are averaged to obtain the final detection result, 
  while standard deviation quantifies detection uncertainty. 
  Notably, regions with high uncertainty often correspond to false alarms.}
  \label{fig:det_uncertainty}
\end{figure}

\textbf{MC Dropout for Detection:}
We apply dropout during inference, generating $N_D$ detection outputs for each frame. We aggregate bounding-box predictions across these $N_D$ forward passes by clustering them based on Intersection-over-Union (IoU) and averaging box parameters within each cluster (Fig.~\ref{fig:det_uncertainty}). We also record the standard deviation of each cluster’s bounding boxes to quantify detection uncertainty. In our experiments, $N_D = 10$.

\subsection{Bayesian Approximation–Based Motion Prediction}
\label{sec:sec3c}
We design a transformer-based motion prediction network \cite{transformer}, which captures the complex temporal dependencies in object trajectories, addressing the shortcomings of simple constant-velocity models. The network takes a tracklet of up to $n$ past states per object as input and outputs the predicted state $\hat{p}_{t+1} = [\hat{x}, \hat{y}, \hat{z}, \hat{\theta}]_{t+1}$.

\begin{figure}[t]
  \centering
  \includegraphics[width=0.65\linewidth]{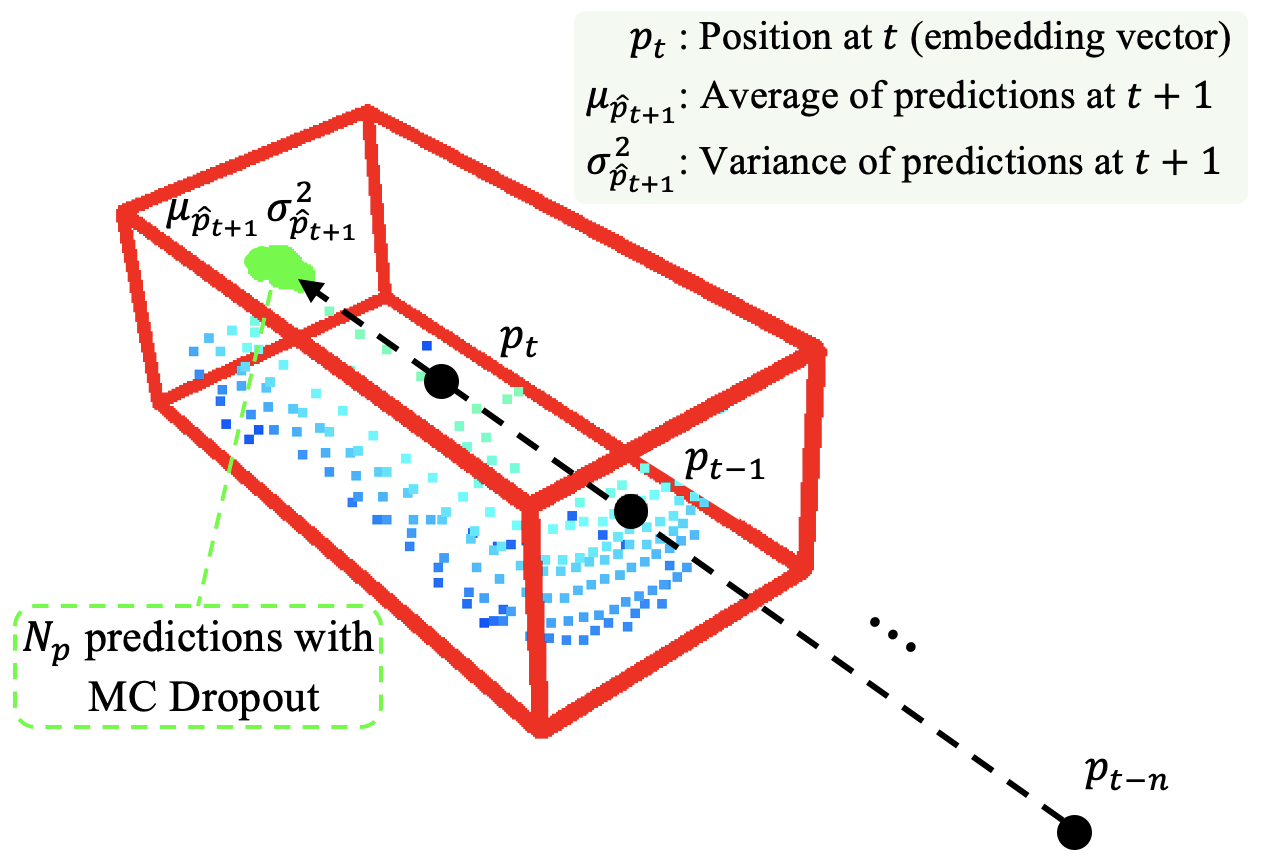}
  \caption{Visualization of MC Dropout–based trajectory predictions.
  Blue dots represent the current object’s 4D Radar measurements, 
  and the red box denotes the object's bounding box at time $t$. 
  The black markers $p_t, p_{t-1}, p_{t-n}$ indicate past object states. 
  Green points show the $N_P$ predicted positions for time $t+1$, 
  whose mean and variance ($\mu_{\hat{p}_{t+1}}, \sigma^2_{\hat{p}_{t+1}}$) 
  capture the estimated trajectory and uncertainty.}
  \label{fig:framework}
\end{figure}

\textbf{MC Dropout for Prediction:}
We again apply dropout at inference, generating $N_P$ trajectory predictions:
\begin{equation}
\mu_{\hat{p}_{t+1}} = \frac{1}{N_P} \sum_{i=1}^{N_P} \hat{p}_{t+1}^i, 
\quad 
\sigma_{\hat{p}_{t+1}}^2 = \frac{1}{N_P} \sum_{i=1}^{N_P} \bigl(\hat{p}_{t+1}^i - \mu_{\hat{p}_{t+1}}\bigr)^2.
\end{equation}
The mean $\mu_{\hat{p}_{t+1}}$ serves as the predicted state, while $\sigma_{\hat{p}_{t+1}}^2$ captures the uncertainty. We set $N_P = 10$ in our experiments.

\begin{table*}[t]
  \centering
  \begin{threeparttable}
    \caption{Performance Comparison of Baseline Models Using Different Input Horizons and Data Association}
    \label{tab:baseline-performance}
    \begin{tabular}{ccccccccc}
    \hline
    \textbf{Model} & \textbf{Input Horizon} & \textbf{Data Association} & \textbf{AMOTA$\uparrow$} & \textbf{AMOTP$\downarrow$} & \textbf{TP$\uparrow$} & \textbf{FP$\downarrow$} & \textbf{FN$\downarrow$} & \textbf{IDS$\downarrow$} \\
    \hline
    \multirow{2}{*}{CV model} 
      & \multirow{2}{*}{---} 
         & Mahalanobis & 0.485 & 1.069 & 9930 & 2329 & 8893 & \textbf{259} \\
      &              & Two-stage & 0.486 & 1.069 & 9939 & \textbf{2314} & 8891 & 261 \\
    \hline
    \multirow{8}{*}{Prediction Network} 
      & \multirow{2}{*}{2} 
         & Mahalanobis & 0.489 & 1.077 & 9945 & 2444 & 8853 & 285 \\
      &              & Two-stage & 0.491 & 1.078 & 9957 & 2430 & 8848 & 285 \\
      & \multirow{2}{*}{3} 
         & Mahalanobis & 0.491 & \textbf{1.042} & 10361 & 2867 & 8406 & 315 \\
      &              & Two-stage & \textbf{0.492} & 1.043 & \textbf{10373} & 2850 & \textbf{8403} & 315 \\
      & \multirow{2}{*}{4} 
         & Mahalanobis & 0.437 & 1.182 & 9473 & 2964 & 9287 & 322 \\
      &              & Two-stage & 0.437 & 1.183 & 9484 & 2950 & 9275 & 323 \\
      & \multirow{2}{*}{5} 
         & Mahalanobis & 0.401 & 1.222 & 9041 & 3112 & 9734 & 307 \\
      &              & Two-stage & 0.402 & 1.222 & 9057 & 3099 & 9725 & 305 \\
    \hline
    \end{tabular}
    \begin{tablenotes}[flushleft]
      \small
      \item $\uparrow$ indicates better performance as the metric value increases.
      \item $\downarrow$ indicates better performance as the metric value decreases.
      \item Bold indicates the best performance in each column.
    \end{tablenotes}
  \end{threeparttable}
\end{table*}

\subsection{Two-Stage Data Association}
\label{sec:sec3d}
We adopt a two-stage data association that first uses Mahalanobis distance and then Doppler-based velocity matching. 

\textbf{Stage 1: Mahalanobis Distance.}
We match detections $z_i$ with predicted states $\hat{z}_j$ by computing:
\[
D_{M}(i, j) = \sqrt{(z_i - \hat{z}_j)^T S^{-1} (z_i - \hat{z}_j)},
\]
where $S$ is the covariance matrix, which we update dynamically based on the prediction’s estimated variance.

\textbf{Stage 2: Relative Velocity.}
Unmatched tracks from Stage 1 are further refined using Doppler velocity. We define a velocity affinity:
\[
A_{R}(i, j) = \exp\Bigl(-\frac{\|\Delta v_{ij}\|^2}{2\sigma^2}\Bigr),
\]
where $\Delta v_{ij}$ is the velocity difference. The final association cost is:
\[
A(i, j) = w_1\, D_{M}(i, j) + w_2\, \bigl(1 - A_{R}(i,j)\bigr).
\]
Objects that remain unmatched after both stages are either initialized as new tracks or removed according to a standard life-cycle management procedure \cite{simpletrack}.

\begin{figure*}[t]
  \centering
  \includegraphics[width=0.8\textwidth]{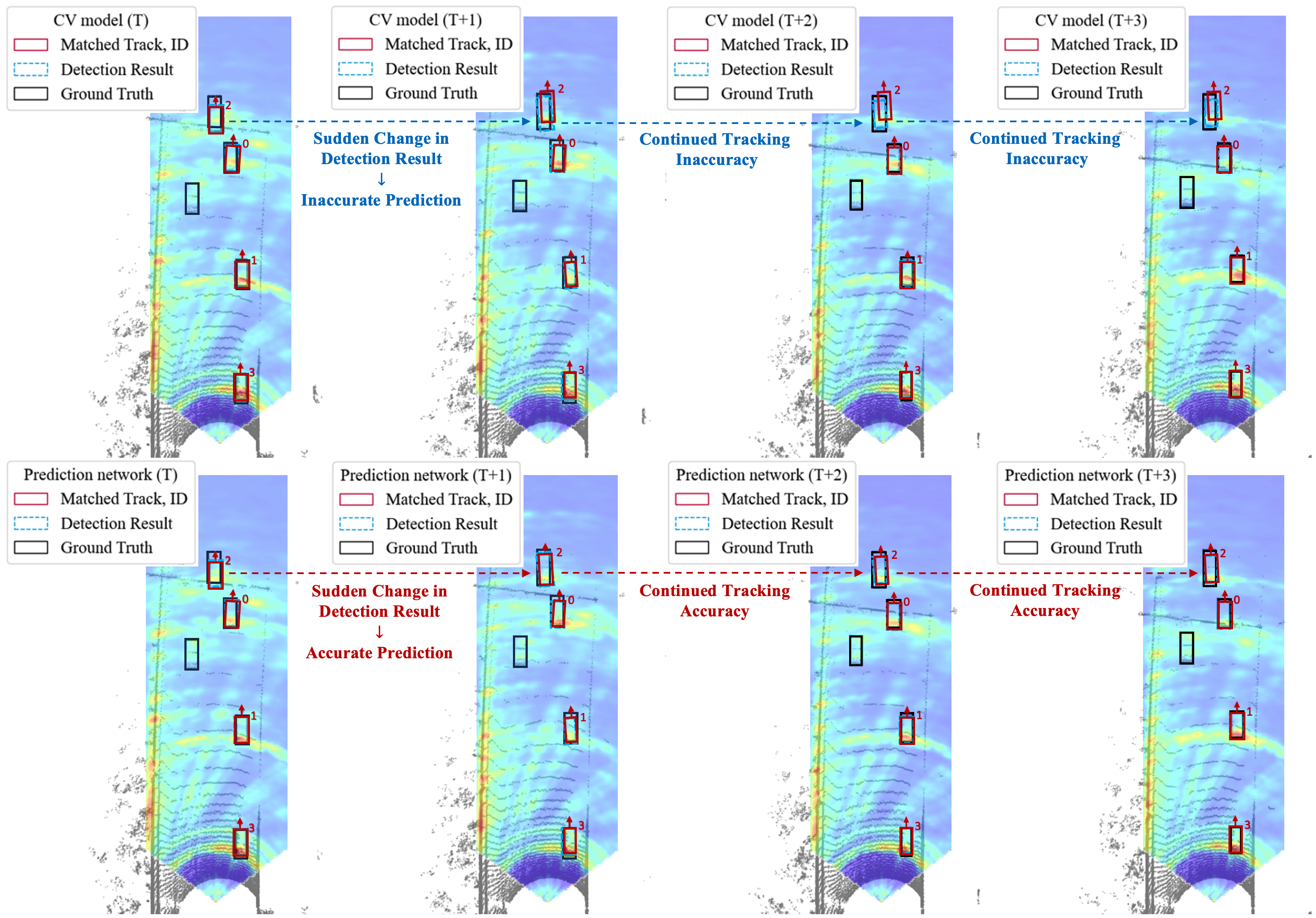}
  \caption{Qualitative comparison of the CV model and the proposed prediction network. Shown are bounding boxes for matched tracks (in red), detection results, and ground truths over four consecutive frames (T, T+1, T+2, T+3) in a bird’s-eye view. Compared to the CV model, which accumulates errors when objects make abrupt changes (e.g., ID 2), the proposed prediction network accurately updates the track over time.}
  \label{fig:qualitative comparison}
\end{figure*}

\begin{table*}[t]
\centering
\caption{Performance Comparison of Various Bayesian Approximation (V denotes using estimated variance.)}
\label{tab:prediction-detection}
\begin{tabular}{cccccccc}
\hline
\textbf{Prediction} & \textbf{Detection} & \textbf{AMOTA$\uparrow$} & \textbf{AMOTP$\downarrow$} & \textbf{TP$\uparrow$} & \textbf{FP$\downarrow$} & \textbf{FN$\downarrow$} & \textbf{IDS$\downarrow$} \\
\hline
Baseline (CV) & RTNH & 0.492 & 1.043 & 10373 & 2850 & 8403 & 315 \\
MC Dropout (Pred) & RTNH & 0.494 & 1.006 & 10351 & 2774 & 8395 & 306 \\
MC Dropout (Pred) + V & RTNH & 0.495 & 1.000 & 10364 & 2774 & 8389 & 299 \\
\hline
Baseline (CV) & Loss Attenuation & 0.515 & 1.011 & 10354 & 2501 & 8538 & 191 \\
MC Dropout (Pred) & Loss Attenuation & 0.519 & 0.946 & 10380 & 2478 & 8526 & 177 \\
MC Dropout (Pred) + V & Loss Attenuation & 0.521 & 0.937 & 10371 & 2483 & 8542 & \textbf{170} \\
\hline
Baseline (CV) & MC Dropout + Loss Attenuation & 0.537 & 0.971 & 11223 & 2969 & 7595 & 265 \\
Baseline (CV) & MC Dropout + Loss Attenuation + V & 0.529 & 1.005 & 11255 & 3175 & 7545 & 283 \\
MC Dropout (Pred) & MC Dropout + Loss Attenuation & 0.539 & 0.912 & 11234 & 2955 & 7627 & 222 \\
MC Dropout (Pred) + V & MC Dropout + Loss Attenuation & \textbf{0.542} & \textbf{0.894} & \textbf{11682} & 3399 & \textbf{7154} & 246 \\
MC Dropout (Pred) & MC Dropout + Loss Attenuation + V & 0.535 & 0.925 & 11236 & 2961 & 7596 & 251 \\
MC Dropout (Pred) + V & MC Dropout + Loss Attenuation + V & 0.537 & 0.915 & 10347 & \textbf{2162} & 8528 & 208 \\
\hline
\end{tabular}
\end{table*}

\section{Experiments}
\label{sec:sec4}
We evaluate Bayes-4DRTrack on the K-Radar dataset \cite{kradar}, focusing on multiple weather conditions (including adverse scenarios) to test robustness. We implement our system on an Intel\textsuperscript{\textregistered} i7-10700 CPU with an NVIDIA RTX 3090 GPU.

\subsection{Dataset and Metrics}
\label{sec:sec4a}
The K-Radar dataset \cite{kradar} contains 4D Radar tensors, LiDAR point clouds, RGB images, RTK-GPS, and IMU data, collected in a wide range of conditions. It provides over 93.3K annotated 3D bounding boxes for vehicles in a 120\,m longitudinal by 80\,m lateral range. This diverse set of scenarios is ideal for evaluating system performance under challenging conditions.

We use standard 3D MOT metrics \cite{tbd}, including Average Multi-Object Tracking Accuracy (AMOTA) and Average Multi-Object Tracking Precision (AMOTP). We also report the number of true positives (TP), false positives (FP), false negatives (FN), and identity switches (IDS). AMOTA and AMOTP aggregate over different recall thresholds and comprehensively capture accuracy and precision, respectively.

\subsection{Ablation Studies and Analysis}
\label{sec:sec4b}

\subsubsection{Comparison of Motion Prediction Network with Constant-Velocity (CV) Model}
We begin by comparing our transformer-based motion prediction network with the classical CV model under varying input horizons ($n=2,3,4,5$). As shown in Table~\ref{tab:baseline-performance}, a horizon of $n=3$ yields the best balance of performance. The CV model underperforms when objects undergo sudden velocity changes, resulting in higher AMOTP. By contrast, our motion prediction network captures nonlinear patterns, producing substantially better AMOTA and AMOTP. A qualitative comparison in Fig.~\ref{fig:qualitative comparison} demonstrates that our model tracks an object undergoing rapid changes more accurately over subsequent frames, whereas the CV model’s errors compound over time.

\begin{table}[t]
\centering
\caption{Performance Comparison with Other 3D MOT Systems \\(Using RTNH for Detection)}
\label{tab:tracking-methods}
\begin{tabular}{cccccccc}
\hline
\textbf{Method} & \textbf{AMOTA$\uparrow$} & \textbf{AMOTP$\downarrow$} & \textbf{TP$\uparrow$} & \textbf{FP$\downarrow$} & \textbf{FN$\downarrow$} & \textbf{IDS$\downarrow$} \\
\hline
\cite{ab3dmot} & 0.339 & 1.281 & 7778 & \textbf{1866} & 10857 & 437 \\
\cite{probabilistic} & 0.485 & 1.069 & 9930 & 2329 & 8893 & 259 \\
\cite{simpletrack} & 0.497 & 1.027 & 10359 & 2651 & 8428 & 295 \\
\textbf{Ours} & \textbf{0.542} & \textbf{0.894} & \textbf{11682} & 3399 & \textbf{7154} & \textbf{246} \\
\hline
\end{tabular}
\end{table}

\subsubsection{Effect of Two-Stage Data Association}
We next demonstrate that including Doppler-based velocity refinement improves the matching of objects with similar spatial characteristics. As shown in Table~\ref{tab:baseline-performance}, the two-stage association consistently increases TP and decreases FP compared to Mahalanobis distance alone, particularly when the input horizon is set to 2, 3, or 5. This confirms that Doppler measurements are valuable for disambiguating challenging cases, leading to overall higher AMOTA.

\subsubsection{Incorporating Bayesian Approximation}
Finally, we perform ablation studies to assess the impact of Bayesian approximation in detection (RTNH + MC Dropout + Loss Attenuation) and prediction (transformer + MC Dropout) under two different configurations of noise covariance in the Kalman filter: fixed variance vs. estimated variance.

\textbf{Fixed Variance vs. Estimated Variance.}  
When using fixed variance in the Kalman filter, applying MC Dropout in the prediction step improves AMOTA from 0.492 to 0.494, while Loss Attenuation in detection raises AMOTA to 0.515 (Table~\ref{tab:prediction-detection}). However, further gains are achieved by using variance derived from MC Dropout as the process noise. Notably, applying the variance from the prediction step (i.e., MC Dropout (Pred) + V) in addition to MC Dropout + Loss Attenuation in detection leads to an AMOTA of 0.542. This highlights that properly modeling uncertainty in the prediction step can meaningfully boost tracking performance.

Interestingly, incorporating the variance from the detection step into the measurement noise sometimes degrades performance (e.g., AMOTA drops from 0.542 to 0.537 in certain settings), suggesting that overemphasizing detection uncertainty can mislead the Kalman filter and reduce effective measurement updates. Overall, results confirm that the best practice in a tracking-by-detection framework is to incorporate adaptive variance primarily in the prediction step, while carefully balancing detection uncertainties.

\subsection{Comparison with Other 3D MOT Systems}
\label{sec:sec4c}
Finally, we compare Bayes-4DRTrack with representative tracking-by-detection systems that use a constant-velocity motion model: AB3DMOT \cite{ab3dmot}, Probabilistic 3D MOT \cite{probabilistic}, and SimpleTrack \cite{simpletrack}. For fairness, all methods use RTNH \cite{kradar} as the base detector. Table~\ref{tab:tracking-methods} shows that our method outperforms the best baseline (SimpleTrack) by 4.5\% in AMOTA, and also achieves the highest TP (11,682) and lowest FN (7,154), as well as the lowest IDS (246). 

Notably, the addition of transformer-based prediction and Bayesian approximation allows our system to maintain accuracy in complex scenarios where other methods deteriorate. By dynamically modeling uncertainty at both detection and prediction stages, Bayes-4DRTrack exhibits improved resilience and consistency, making it well-suited for real-world autonomous driving tasks.

\section{Conclusion}
\label{sec:sec5}
This paper presents \emph{Bayes-4DRTrack}, a novel 3D MOT system for 4D Radar that integrates Bayesian approximation in both detection and prediction. By leveraging a transformer-based motion prediction network and Doppler-based two-stage data association, the system addressed abrupt motion changes and challenging tracking scenarios with improved accuracy. Experimental results on the K-Radar dataset showed that our method outperforms state-of-the-art systems, offering a significant boost in AMOTA (by 5.7\%). These findings highlight the importance of dynamic uncertainty modeling, which helps maintain robust performance under real-world conditions where classical fixed-noise models often fail.

\section*{Acknowledgement}
This work was supported by the National Research Foundation of Korea(NRF) grant funded by the Korea government(MSIT) (No. 2021R1A2C3008370).

\bibliographystyle{IEEEtran}
\bibliography{egbib}

\end{document}